\documentclass{article}

\usepackage{microtype}
\usepackage{graphicx}
\usepackage{subfigure}
\usepackage{booktabs} 

\usepackage[accepted]{icml2019}
\usepackage{seb}

\icmltitlerunning{Towards Robust Evaluations of Continual Learning}

\begin{document}

\twocolumn[
\icmltitle{Towards Robust Evaluations of Continual Learning}



\icmlsetsymbol{equal}{*}

\begin{icmlauthorlist}
\icmlauthor{Sebastian Farquhar}{ox}
\icmlauthor{Yarin Gal}{ox}
\end{icmlauthorlist}

\icmlaffiliation{ox}{OATML Research Group, Department of Computer Science, University of Oxford, United Kingdom}

\icmlcorrespondingauthor{Sebastian Farquhar}{sebastian.farquhar@cs.ox.ac.uk}

\icmlkeywords{Machine Learning, ICML, Continual Learning, Incremental Learning, Sequential Learning, Variational Inference, Catastrophic Forgetting, Deep Neural Networks}

\vskip 0.3in
]



\printAffiliationsAndNotice{}  

\begin{abstract}
Experiments used in current continual learning research do not faithfully assess fundamental challenges of learning continually. Instead of assessing performance on challenging and representative experiment designs, recent research has focused on increased dataset difficulty, while still using flawed experiment set-ups. We examine standard evaluations and show why these evaluations make some continual learning approaches look better than they are. We introduce desiderata for continual learning evaluations and explain why their absence creates misleading comparisons. Based on our desiderata we then propose new experiment designs which we demonstrate with various continual learning approaches and datasets. Our analysis calls for a reprioritization of research effort by the community.
\end{abstract}
\section{Introduction}
\label{introduction}
Certain applications require \textit{continual} learning---splitting training into a series of tasks and discarding the data after training each task. These settings vary tremendously. If data relate to individuals it may be unethical, illegal, or imprudent to retain old datasets: for example, a hospital might want to delete old patient data. In other applications, real-time systems could face distributional shifts with an underlying distribution which changes faster than the time it would take to retrain a new model with all the data. For example, a Mars rover might cross regions of different terrain that require adaptation without forgetting. Neural networks trained on such series of tasks tend to forget earlier tasks (often referred to as \textit{catastrophic forgetting}).

Recent works have shown promising advances towards continual learning by adding regularization terms to the loss function which preserve important parameters \citep{kirkpatrick_overcoming_2017, Zenke2017, nguyen_variational_2018, Chaudhry2018, Ritter2018}. We call these approaches \scare{prior-focused} because they can be interpreted as treating intermediate models as priors when learning updated weights. Although a useful starting point, experimental evaluations in these recent works have major blind-spots which mask weak-points of current approaches. The design of the evaluations does not fully reflect the core motivations for continual learning, regardless of the dataset used, including evaluations based on MNIST \citep{LeCun1998}, notMNIST \citep{Bulatov2011}, FashionMNIST \citep{xiao_fashion-mnist:_2017}, CIFAR10 \citep{Krizhevsky2009} and others. But evaluations which obscure the shortcomings of suggested continual learning solutions impede developments in the field, since researchers are not made aware of limitations of past research. 

In this paper, we explore desiderata for continual learning evaluations based on real-world uses. Applications of continual learning are very diverse, and authors consider a wide range of settings. As a result, rather than propose a single benchmark which risks being overly narrow, we propose fundamental desiderata for empirical evaluations in continual learning to make them more representative of the challenges that motivate the field. We then demonstrate that experiments which neglect our core desiderata can be misleading---a continual learning system that performs well when a subset of the desiderata are in place might fail entirely when a core desideratum is imposed. No prior-focused continual learning system has so far been shown to succeed when all five desiderata are applied. We demonstrate that several leading prior-focused approaches stop working on a more robust evaluation. We therefore argue that existing evaluations are biased, because prior-focused approaches appear to succeed, but have major blind-spots. Lastly, we introduce new evaluations based on our list of desiderata which offer richer challenges for continual learning systems.

Specifically, our four main contributions are as follows:
\begin{enumerate}
\item We propose fundamental desiderata for future evaluations, which can be applied regardless of dataset.
\item We analyse the shortcomings of existing widely used evaluations in continual learning.
\item We show empirically that existing evaluations are biased towards prior-focused approaches.
\item We propose new experimental designs which mitigate the issues of existing ones.
\end{enumerate}

\section{Continual Learning}
We start by formalising the definition of \textit{continual learning}.
In a typical supervised learning setting, we aim to learn parameters $\mathbf{w}$ using an independently and identically distributed (i.i.d.) labelled training dataset $\mathcal{D} \equiv \{(\mathbf{x}^{(i)}, y^{(i)})\}$ to accurately predict $p(y^*|\mathbf{w}, \mathbf{x}^*)$ for an unseen $(\mathbf{x}^*, y^*)$ pair.

In the continual learning setting, members of $\mathcal{D}$ are not i.i.d. Instead they may be split into disjoint subsets $\mathcal{D}_{t} \equiv \{(\mathrm{x}_t^{(i)}, y_t^{(i)})\}$. These sets are assumed to be drawn from $T$ distinct i.i.d. distributions each of which represents a \textit{task}. The challenge of continual learning is to learn a single model which is able to predict well on data from any task, despite training on each task in sequence without deliberately revisiting previous tasks. For some applications, stronger or weaker assumptions are made. For example, \citet{nguyen_variational_2018} concern themselves with fast adaptation, so they allow themselves to retain a small coreset of data from old tasks. Similarly, \citet{kirkpatrick_overcoming_2017, schwarz_progress_2018} consider a simulation in which they allow themselves to revisit each task multiple times. In this paper, we will frequently consider a model after training on some, but not necessarily all $T$ tasks. The $t$'th model is the state of the model after training on datasets $\{D_1:D_t\}$.

\section{Proposed Desiderata for Continual Learning Evaluations}\label{desiderata}
The setting and motivation for continual learning can vary enormously, from image segmentation to medical diagnostics to control. Specialized continual learning datasets like Core50 \citep{Lomonaco2017} or environments such as that in \citet{schwarz_towards_2018} are therefore useful for subfields but cannot be seen as an overall benchmark for continual learning. We therefore concentrate instead on developing set of sufficient desiderata for future evaluations of continual learning. We propose the following fundamental principles which ought either to be respected or to be explicitly mentioned as absent. In \S\ref{critical_analysis} we offer a critical analysis of existing experiments which suggests why core desiderata are important. In \S\ref{failure_modes}, we present a set of experiments showing how neglecting these core aspects can make a continual learning system look more successful than it is.

Use cases of continual leaning vary tremendously. One example is a hospital developing a system for automated disease diagnosis using patient sub-population A which is then refined in a different hospital with patient sub-populations B (with sub-population A data not allowed to leave the first hospital). A second example is a wind-turbine safety system which predicts wind velocity an hour ahead to decide whether to turn off the turbine in case of strong winds, with wind dynamics changing between the seasons of the year. A third example is an ad serving system that has to update with huge amounts of streaming data, without feasibly being able to keep all data around. Yet another use case is a Mars rover updating its behaviour as it encounters new types of terrain---perhaps much looser or coarser soil than it has seen before---but which must remain able to manoeuvre on old terrain. All use cases share common core desiderata, which we demonstrate next using the Mars rover example for concreteness. Note that this example is for illustrative purposes only---we are \textit{not} proposing that this example should be used for evaluations. Our desiderata arise just as readily from other settings.

\begin{compactenum}
\item[\textbf{A: Cross-task resemblances}]{Input data from later tasks must resemble old tasks enough that they at least sometimes result in confident predictions of old classes, early in training. The widely used Permuted MNIST (see \S\ref{permuted_description}) which violates this would correspond to every input sensor in our Mars rover being randomly rewired---unlikely to be representative of real cases.}
\item[\textbf{B: Shared output head}]{If each task is given a different output vector during training and testing, it must be explicitly and prominently mentioned. This has a large effect on the difficulty of the challenge. In our example, the rover is using the same outputs to control its motors in all terrain.}
\item[\textbf{C: No test-time assumed task labels}]{Related to desideratum B, if we knew in advance what the tasks were and had a way to distinguish them, one could have a different model for each task and switch between them. In our example, if the rover has a good way to distinguish rock from dust and can learn a separate policy for each it might not need continual learning.}
\item[\textbf{D: No unconstrained retraining on old tasks}]{Many motivations of continual learning preclude being able to retrain on task $1$ after having trained on task $T$. In our example, the rover needs to actually move into terrain it has left behind before it can train there. In other examples, even retaining small amounts of data from old tasks might violate privacy laws}
\item[\textbf{E: More than two tasks}]{The more tasks a continual learning system can handle the better. Our rover, for example, may face a potentially open-ended and large number of tasks. Two-task transfer (\S\ref{two-task}) is therefore interesting, but succeeding at two-task transfer does not guarantee good performance on more tasks.}
\end{compactenum}

In certain applications we might deliberately ignore some of the desiderata. For example, \citet{nguyen_variational_2018} neglect desiderata B and C because they consider fast adaptation to new tasks which have labels during both training and testing. In a setting with easy access to a simulation of all tasks, retraining might be permissible, setting aside desideratum D (as in \citet{schwarz_progress_2018}). Desideratum E is mostly set aside for convenience---there might be interesting intermediate research which is nevertheless useful progress for the field. Desideratum A is mostly neglected for historical reasons discussed in \S\ref{crit_permuted}. While the five listed desiderata will not always be applicable, it remains a surprising fact that there is \textbf{no prior-focused approach that has been shown to perform well in an evaluation with all of the five core desiderata}.

The core desiderata are a starting point which can lead to more difficult challenges, some of which are beginning to be addressed by recent work, including:
\begin{compactenum}
\item[\textbf{Unclear task demarcation}]{Task boundaries are generally assumed knowledge during training.}
\item[\textbf{Continuous tasks}]{Current practice usually assumes that tasks are discretely different rather than varying continuously.}
\item[\textbf{Overlapping tasks}]{Current practice tends to have tasks where each task has its own classes which are disjoint from the other tasks. \citet{Lomonaco2017} discuss this point.}
\item[\textbf{Long task sequences}]{Continual learning is most interesting over extremely long sequences of tasks.}
\item[\textbf{Time/compute/memory constraints}]{It is widely acknowledged that continual learning solutions that update quickly or use small/fixed memory are more useful than ones that have large computational overheads.}
\item[\textbf{Strict privacy guarantees}]{For settings motivated by privacy, tracking the differential privacy guarantees between tasks is important.}
\end{compactenum}

Each of these further desiderata offers a way to make a continual learning evaluation more representative of the needs of applications \textit{without making the dataset itself harder}. In \S\ref{further_evaluations} we suggest two experiments that address the most interesting of these. In \S\ref{mutual_information} we show how model uncertainty can be used to detect task changes, which is essential in many practical settings and allows streaming settings to be converted into continual learning ones. In \S\ref{time} we show how training time and accuracy must be traded off against each other as design choices, rather than just reporting the time taken by the most accurate system. Memory can be treated like time, here, and the other desiderata are straightforward to construct from existing datasets by tweaking the dataloader.

We note that a very real risk for the field is to \textit{embrace more complex datasets before establishing robust experimental design.} Until we have continual learning evaluations that test the core challenges of continual learning, moving from MNIST to more complex datasets does little to measure progress in continual learning itself.
\begin{figure}
    \centering
    \includegraphics[width=0.8\columnwidth]{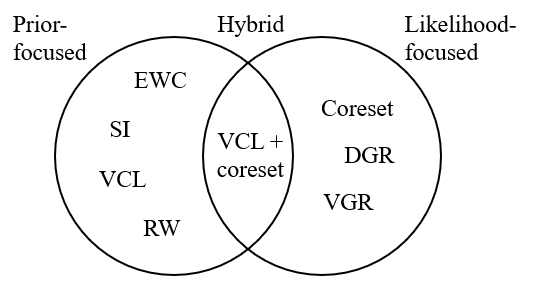}
    \vspace{-5mm}
    \caption{We contrast prior-focused with likelihood-focused continual learning. By comparing these, and hybrid forms, we can see which evaluations pose a bigger challenge to different approaches.}
    \label{fig:venn}
    \vspace{-3mm}
\end{figure}

\section{Existing Work} \label{prior_work}
In this paper, we critically analyse common evaluations used by the majority of recent papers. In this section, we review existing work focusing first on the methods employed (\S\ref{methods}) and second on the evaluations used (\S\ref{evaluations}). Throughout, we provide a framing which anticipates our critical analysis.

\subsection{Methods}\label{methods}
Approaches to continual learning and catastrophic forgetting form three main families. First, prior-focused approaches use regularization to create \scare{elastic} parameters. These include: Elastic Weight Consolidation (EWC) \citep{kirkpatrick_overcoming_2017, huszar_note_2018}, Synaptic Intelligence (SI) \citep{Zenke2017}, Variational Continual Learning (VCL) \citep{nguyen_variational_2018}, Kronecker-factored approximated Laplace approximation \citep{Ritter2018}, and Riemanian Walk (RW) \citep{Chaudhry2018}. In fact, all of these can be seen as setting a distribution over the parameters of the $t-1$'th model as the prior when training on the $t$'th dataset.

VCL makes this most explicit. VCL is a variational inference (VI) method that uses Bayesian neural networks \citep{MacKay1992, Neal1995, Blundell2015}. VCL sets the posterior at the end of training a task to be the prior when beginning training for the next task. Schematically, as the authors of these papers note, all the approximately Bayesian approaches use a similar loss function:
\begin{equation}
\label{eq:prior}
\begin{split}
    \mathcal{L}_t = &\sum_i \mathrm{log}\big(p(y_t^{(i)}|\mathbf{w}_t, \mathbf{x}_t^{(i)})\big) \\
    &- \frac{c}{2}(\mathbf{w}_t - \mathbf{w}_{t-1})^T\Sigma_{t-1}^{-1}(\mathbf{w}_t - \mathbf{w}_{t-1})
\end{split}
\end{equation}
where $\Sigma$ is the covariance of the prior, which is estimated differently under each approach. The first term reflects model log-likelihood on the $t$'th task. The second reflects distance from the model prior.

Prior-focused approaches rely on the prior term to capture everything learned on previous tasks and only use the newest data to estimate the likelihood term. However, in practice the prior term is only approximated. \citet{farquhar_unifying_2018} contrast prior-focused approaches with a second family, likelihood-focused approaches, that attempt to estimate the likelihood of the current model on past tasks  (see figure \ref{fig:venn}). Examples of this include Deep Generative Replay (DGR) by \citet{Shin2017}, which uses generative models to store summaries of old tasks, or Variational Generative Replay (VGR) a variational and simplified alternative presented in \citet{farquhar_unifying_2018}. Simply storing a small random sample or coreset is an alternative approach in this family. Many members of this family are called rehearsal strategies after `pseudorehearsal' introduced by \citet{Robins1995}, or called dual-memory strategies following a neuro-scientific motivation. However, some dual-memory strategies, e.g., GEM \citep{lopez-paz_gradient_2017} or model adaptation \citep{hu_overcoming_2019}, are not likelihood-focused.

Some authors have observed better results for prior-focused methods with the addition of some likelihood estimation, although this has not been rigorously explored. For example, VCL is also presented using coresets---fine-tuning on a small withheld sample just before testing. Similarly, RW presents results both with and without training on subsamples of previous datasets.

The last major family uses dynamic architectures. These change the structure of the networks in significant ways to incorporate learning from each task separately (e.g., \citet{Rusu2016, Li2017}). These perform well, but in most cases their need for new models for each task creates growing compute/memory costs that do not seem to truly solve the continual learning problem. A more thorough survey of these as well as prior- and likelihood-focused approaches can be found in Appendix \ref{furtherpriorwork}. We focus our examination on prior-focused and their counterpart likelihood-focused methods, because they seem to represent the most exciting recent advances in continual learning, but the experiments used to support them may not be fairly evaluating their performance.

\subsection{Experimental Evaluations}\label{evaluations}
In this section we give an in-depth review of key experimental evaluations as they are used in \textit{current research}. We illustrate our critique using the most commonly performed experiments using variations of the MNIST dataset: \textit{Permuted MNIST} and \textit{Split MNIST}, most commonly in \textit{multi-headed} form. Although we emphasise MNIST for explanatory purposes, our critique in \S\ref{critical_analysis} applies to many other evaluations which share similar design choices.

\subsubsection{Permuted MNIST}\label{permuted_description}
The Permuted MNIST experiment was introduced in \citet{Goodfellow2013}. In their experiment, a model is trained on MNIST as $\mathcal{D}_1$. Each later $\mathcal{D}_t$ for $1 < t \leq 10$ is constructed from the MNIST data but with the pixels of each digit randomly permuted. A fresh permutation is drawn for each task and applied to all images in the same way for that task. After training on each dataset, one evaluates the model on each of the previous datasets as well as the current. \citet{Goodfellow2013} used this experiment to investigate feature extraction, but it has since become a mainstay for continual learning evaluation \citep{Zenke2017, Shin2017, kirkpatrick_overcoming_2017, Lee2017, lopez-paz_gradient_2017, nguyen_variational_2018, Ritter2018, hu_overcoming_2019, chaudhry_efficient_2019}.

\subsubsection{Split MNIST}\label{prior_split}
The Split MNIST experiment was introduced by \citet{Zenke2017} in a \textit{multi-headed} form and used by other authors including \citet{Shin2017, nguyen_variational_2018, Chaudhry2018} (\citet{Ritter2018} use a two-task variant). The experiment constructs a series of five related tasks. The first task is to distinguish the digits (0, 1), then (2, 3) etc. Most papers use a multi-headed variant the model prediction is constrained to be only from the two classes represented in each task. For example, when evaluating the performance on the first task, the model only needs to predict probabilities for zero versus one. In some cases, multi-heading is taken even further and training is only done on the head governing the specific classes included in the task \citep{Zenke2017, nguyen_variational_2018, Ritter2018}. A \textit{single-headed} version does not limit predictions during either training or testing. As \citet{Chaudhry2018} note, the multi-headed variant is much easier to solve. Multi-heading is often used in similar non-MNIST evaluations, for example, in \citet{Zenke2017, nguyen_variational_2018, Ritter2018}. \citet{Chaudhry2018} use both a single- and multi-headed version of Split MNIST. \citet{hu_overcoming_2019} use a single-headed set-up but their method effectively learns a different head for each task.

\subsubsection{Two-task transfer}\label{two-task}
In some works, a two-task transfer learning evaluation is used for continual learning. For example, \citet{Shin2017} train first on MNIST and then on SVHN \citep{Netzer2011} (or vice versa) in order to see whether their algorithm preserves performance on the first task after training on the second. \citet{Jung2016a}, \citet{Li2017}, and \citet{pfulb_comprehensive_2018} perform similar experiments.

\subsubsection{Desiderata, benchmarking and metrics}
Indicating the importance of robust evaluations, several authors have recently included some continual learning desiderata alongside their work. For example, \citet{schwarz_progress_2018} list desiderata for continual learning systems which includes being able to handle long sequences of tasks without knowing task labels and ideally no clear task demarcation, but do not introduce experimental set-ups which could enforce all these desiderata, such as the ones we propose in \S\ref{further_evaluations}. \cite{chaudhry_efficient_2019} call for time and memory constraints as well as careful use of cross-validation. \citet{pfulb_comprehensive_2018} call for testing on many datasets, care with cross-validation, and storage constraints. \citet{farquhar_differentially_2018} discuss the importance of tracking differential privacy guarantees in continual learning settings motivated by privacy.

Several authors have commented on the importance of systematic benchmarks for continual learning, including \citet{Lomonaco2017}, focused on object recognition, and \citet{schwarz_towards_2018}, focused on control in a StarCraft environment. Others have worked towards richer metrics for measuring continual learning performance like forward/backwards transfer or learning curve area \citep{lopez-paz_gradient_2017, Chaudhry2018, schwarz_progress_2018, diaz-rodriguez_dont_2018, chaudhry_efficient_2019}. Both of these developments are orthogonal to the arguments of this paper --- well-chosen metrics evaluated in a challenging environment are still unhelpful if the experiment is not designed to reflect the continual learning needs of diverse applications.

\section{Critical Analysis of Existing Evaluations}\label{critical_analysis}
Pixel permutation (e.g., Permuted MNIST), split classification (e.g., multi-headed Split MNIST), and two-task transfer experiments are commonly used to compare continual learning algorithms. Although they are a useful starting point, each of these makes continual learning easier for prior-focused approaches. All of our critiques of the way MNIST is used by \citet{Zenke2017} apply also to their use of the CIFAR100 dataset, the use of notMNIST, Fashion MNIST, SVHN and CIFAR10 in \citet{Ritter2018}, and the use of notMNIST in \citep{nguyen_variational_2018}. These critiques are about experimental design, not dataset choice.

\subsection{Pixel Permutation}\label{crit_permuted}
Permuted MNIST represents an unrealistic best case scenario for continual learning---although it satisfies the literal definition of continual learning. The positions of the pixels are fully randomized---which suited its original purpose. \citet{goodfellow_empirical_2013} investigated whether neural networks have `high-level concepts' that get re-mapped to pixel positions when the input space is permuted, and found evidence they did not. Now, however, the experiment has been repurposed for continual learning, where it is not suitable.

An image from each permuted dataset is practically unrecognizable given previous datasets. The actual world is almost never structured like this---new situations look confusingly similar to old ones---the sensor input in a Mars rover will never be permuted no matter what terrain one moves onto. A model presented with a new task in Permuted MNIST will be uncertain, while in settings that are not deliberately randomized a model will tend to make confident but false predictions. This makes Permuted MNIST significantly different from real-world settings in a way that directly affects how new tasks are learned. In Appendix \ref{permuted_problems}, we offer an empirical investigation of this phenomenon and a hypothesis for the mechanism by which Permuted MNIST simplifies the continual learning challenge.

\subsection{Split Classification}\label{crit_split}
The multi-headed version of Split MNIST, which is most often used, is easier but less relevant to applications because it requires knowledge of the task and the classes represented in each task, as \citet{Chaudhry2018} point out. Usually, if that were possible, continual learning would be unnecessary, since one could use a separate classifier for each task, which generally would perform better.

Unfortunately, prior-focused continual learning systems tend to look much better in multi-headed evaluations than single-headed ones. Suppose some model has already been trained on the first four tasks of Split MNIST and is then tested on the digit \scare{1}. In the single-headed variant, when shown a \scare{1} it may incorrectly predict the label is seven, which was seen more recently. In the multi-headed variant, we \textit{knowingly assume} that the label comes from [0:1]. Because the model only needs to decide between 0 and 1 (and not even consider if the image is a 7), a multi-headed model could correctly predict the label is 1 even though the same approach will completely fail in a single-headed experiment setup. Most current prior-focused continual learning systems are good enough that models retain the ability to distinguish old classes, but only if they are able to know which task the example is from. A multi-headed evaluation can therefore make it seem as if an approach has solved a continual learning problem when it has not. \textit{Single-headed Split MNIST}---a less-used evaluation---is the simplest experimental setting that meets all five of our core desiderata. We recommend it as a toy baseline, which is easily adapted to harder datasets. It satisfies our core desiderata: \textbf{A:} The tasks resemble each other---e.g., a seven can look like a one; \textbf{B:} Single-heading means all outputs are shared; \textbf{C:} A full prediction is made over all possible outputs each time; \textbf{D:} Each task is trained only once and only the model is carried forwards; and \textbf{E:} There are five tasks.

\subsection{Other non-representative evaluations}
Two-task transfer is not representative of realistic problems because continual learning use-cases might require a long series of tasks, not just two. An algorithm might have elements that perform well with just one previous task but fail with more. For example, the approximate Fisher information matrix used in EWC is estimated using a Taylor expansion which is only locally accurate in one part of parameter-space. If, after many tasks, the model reaches a very different part of parameter-space, the estimated Fisher information for old tasks will be inaccurate. Moreover, in practice two-task transfer is just easier than true continual learning: \citet{pfulb_comprehensive_2018} conclude that a simple fully connected network will show no catastrophic forgetting on a two-task Permuted MNIST but do not consider longer task sequences, while other authors have already shown that with more tasks there is indeed catastrophic forgetting \citep{kirkpatrick_overcoming_2017}.

\section{Empirical Analysis of Existing Evaluations}\label{empirical_desiderata}
This section has three aims. First, we show that experiments which satisfy only a subset of the five main desiderata from \S\ref{desiderata} can have blind-spots that are important to applications that demand all five desiderata. We do this by showing that leading methods fail on an experiment that uses all five desiderata, while they look good on experiments that use just a subset. Second, we show that existing evaluations are biased towards prior-focused approaches. Third, we suggest two further experimental set-ups.

\subsection{Experimental approach}
We consider representatives of the prior-focused approach, likelihood-focused approach, and a hybrid approach. We use these to demonstrate the shortcomings of existing evaluations and to tease out which aspects of the architecture are challenged under different experimental design choices.

We use three variants of VCL because it has the clearest Bayesian interpretation. First, we use pure VCL without a coreset, exactly as described by \citet{nguyen_variational_2018} (see Appendix \ref{furtherpriorwork}). Second, we use a small coreset of 40 datapoints extracted from each dataset (we use their k-center coreset approach, but this does not have a large effect). The second method is the same as pure VCL except that, at the end of training on each task, the model is trained on the coresets before testing, as described in their work. This reflects a hybrid approach. Third, we use a \scare{coreset only} approach as a likelihood-focused version. It is exactly like the second variant except that the prior used for variational inference is the initial prior each time---it is not updated after each task.\footnote{This is not the coreset only algorithm used in \citet{nguyen_variational_2018}. Theirs is seeing \textit{only} coresets---much less data---which is why it performs badly on even the first task.}

We use two further baselines to show that the effects we find are not an artefact of VCL specifically. EWC \citep{kirkpatrick_overcoming_2017} is used as a prior-focused alternative to VCL because it has been widely discussed in the literature. VGR \citep{farquhar_unifying_2018} is used as a more powerful likelihood-focused alternative to coresets. VGR, VCL and its variants use a Bayesian neural network (BNN) and variational inference, while EWC does not, so performance comparisons between these architectures should be interpreted carefully. The fact that our observations for VCL and its variants are mirrored for EWC supports our hypothesis, based on our analysis, that the problems discussed in this section affect prior-focused approaches in general.
\begin{figure}[t]
\vspace{-2mm}
\includegraphics[width=\columnwidth]{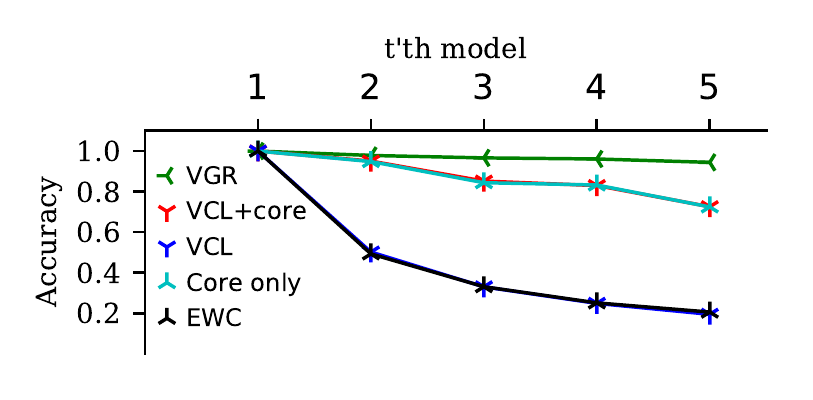}
\vspace{-11mm}
\caption{\textbf{Single-headed Split MNIST.} This experiment meets all core desiderata and shows big performance differences.}
\label{fig:split_single_all}
\end{figure}

\begin{figure}[t]
\vspace{-5mm}
\includegraphics[width=\columnwidth]{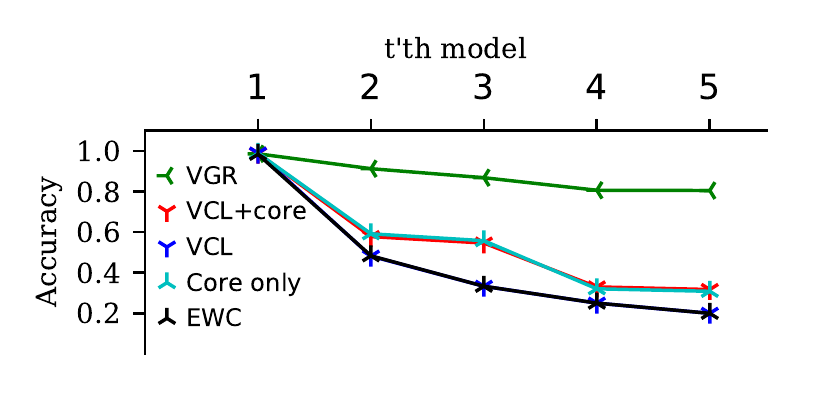}
\vspace{-11mm}
\caption{\textbf{Single-headed Split Fashion MNIST.} The harder dataset shows the prior-approximation starting to deteriorate, but does not change performance ranking.}
\vspace{-2mm}
\label{fig:fashion_single}
\end{figure}

\subsection{Experiments With All Five Desiderata Show Big Performance Differences}
We use Single-headed Split MNIST, described in \S\ref{crit_split}. This satisfies all main desiderata from \S\ref{desiderata}. It reveals major differences in performance between the approaches (see figures \ref{fig:split_single_all} and \ref{fig:fashion_single}). The performance of VCL with coreset appears to be entirely driven by the presence of the coresets. When coresets are removed, VCL alone completely forgets old tasks---its accuracy comes from correctly classifying the most recent task only. EWC performs exactly like VCL, suggesting that it is prior-focused approaches in general that are struggling. Meanwhile VGR, the more accurate likelihood-focused approach, is the only method that performs nearly perfectly. Using FashionMNIST rather than MNIST is harder and a worthwhile additional test, but does not reveal a radically different story. VCL performs much worse on FashionMNIST than VGR even though both use the same model. It seems the prior approximation struggles with data complexity more than the GAN.
\begin{figure}[t]
\centering
\begin{minipage}{\columnwidth}
\vspace{-2mm}
\includegraphics[width=\columnwidth]{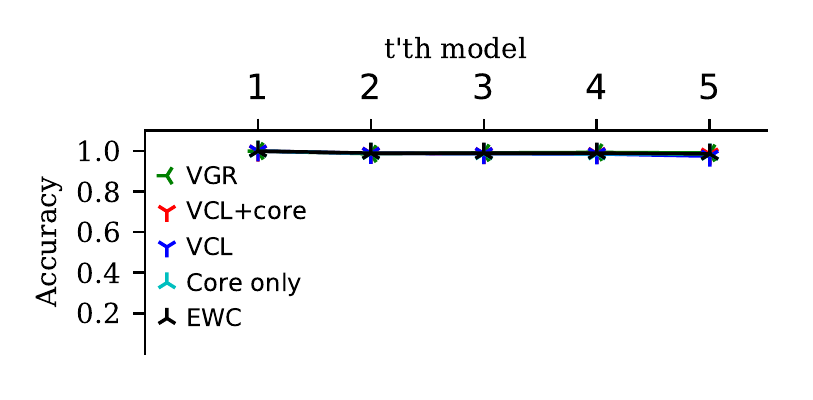}
\vspace{-12mm}
\caption{\textbf{Multi-headed Split MNIST.} All methods succeed.}
\label{fig:split_multi_all}
\end{minipage}
\begin{minipage}{\columnwidth}
\includegraphics[width=\columnwidth]{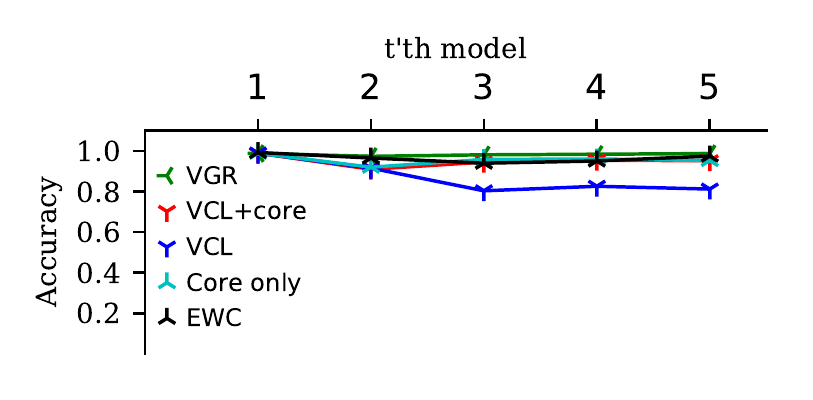}
\vspace{-12mm}
\caption{\textbf{Multi-headed Split FashionMNIST.} All perform similarly, so no clean differentiation. VCL performs slightly worse without coresets.}
\label{fig:fashion_multi}
\vspace{-6mm}
\end{minipage}
\end{figure}

\subsection{Missing Desiderata Create Blindspots}\label{failure_modes}
Having shown that there are big differences in performance between our baselines on a well-constructed experiment that follows all desiderata, we now show that experiments that do not include all desiderata can fail to reveal these important performance differences. This means that these experimental evaluations can be misleading. The fact that they systematically hide shortcomings of prior-focused approaches suggests that the evaluations are inadvertently biasing research efforts in the field. We use MNIST and FashionMNIST to show that while using more complex datasets does improve evaluations, it does not make up for unrepresentative experimental design.

\subsubsection{No cross-task resemblance}
As we argue in \S\ref{crit_permuted}, Permuted MNIST is unusual in that there is no resemblance between images in each task that might cause confident incorrect predictions. Previous authors have demonstrated the effectiveness of their methods on Permuted MNIST \cite{kirkpatrick_overcoming_2017, nguyen_variational_2018}. Virtually any continual learning method proposed has shown good performance on this task. This demonstrates the weakness of experiments that do not reflect desideratum A. Further experimental investigation of these phenomena is in Appendix \ref{permuted_problems} with baseline results in Appendix \ref{perm_details}.

\subsubsection{No shared output head}
Authors frequently use a multi-headed version of Split MNIST, following \citet{Zenke2017}, which we describe in \S\ref{prior_split}. On this evaluation, all approaches look similarly good (see figures \ref{fig:split_multi_all} and \ref{fig:fashion_multi}). This shows that multi-heading, and neglecting desideratum B, can create misleading experiments. Details of training and hyperparameters can be found in Appendix \ref{multi_details}.

\begin{figure}[t]
\centering
\begin{minipage}{\columnwidth}
\vspace{-2mm}
\includegraphics[width=\columnwidth]{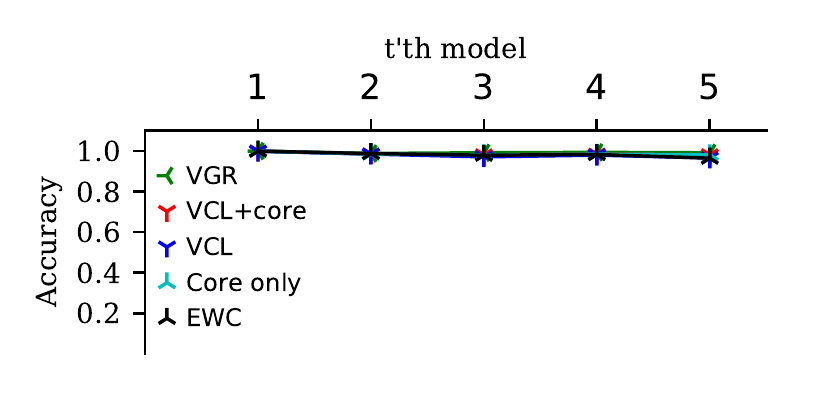}
\vspace{-12mm}
\caption{\textbf{Test-time Knowledge Split MNIST.} All succeed.}
\label{fig:ttk_mnist}
\end{minipage}
\begin{minipage}{\columnwidth}
\includegraphics[width=\columnwidth]{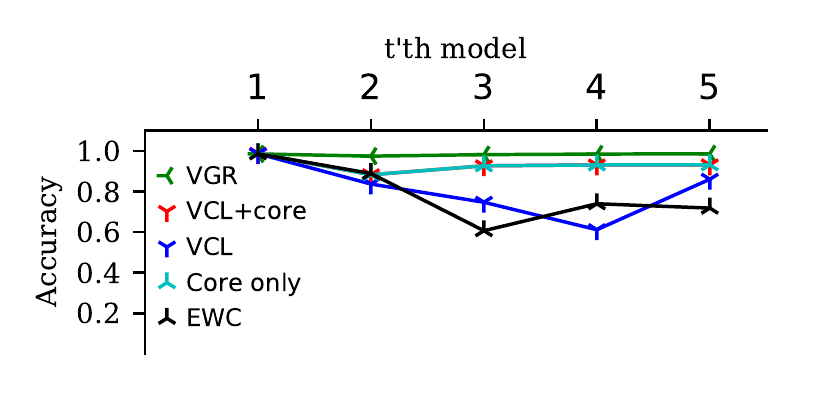}
\vspace{-12mm}
\caption{\textbf{Test-time Knowledge Split FashionMNIST.} The experiment is harder but still does not cleanly differentiate prior- and likelihood-focused methods.}
\label{fig:ttk_fashion}
\end{minipage}
\vspace{-6mm}
\end{figure}

\subsubsection{Test-time Task Knowledge}
Instead of entirely separate heads for each task, one might simply know which output classes to look at during testing. In this variant, which breaks desideratum C, everything is exactly the same as for the single-headed variant during training, but during testing all outputs except those for the correct class are masked with zeros before the most probable class is selected. Here, both on MNIST and FashionMNIST the evaluation does not differentiate the performance of prior- and likelihood-focused methods (see figures \ref{fig:ttk_mnist} and \ref{fig:ttk_fashion}).

\subsubsection{Retraining on old tasks}
An experimental evaluation that is exactly like single-headed Split MNIST except that the tasks are looped through several times does still find that prior-focused methods perform badly (see Appendix \ref{retraining}). Neglecting this desideratum, therefore, might not make evaluations biased, though it is certainly not in keeping with the core motivations of continual learning applications since it requires data to be kept indefinitely.

\subsubsection{Only two tasks}
Many authors have tested their methods with only two tasks. This is an interesting challenge, but significantly simpler than a longer series. Showing that a method succeeds on a two-task transfer does not entail that it will work on a longer series. For example, looking at the first two tasks in figure \ref{fig:split_single_all} would not differentiate VCL with coresets and VGR, where looking at the longer series shows a clear difference in performance.

\subsection{Further Experimental Options}\label{further_evaluations}
Having shown the importance of our five main desiderata in experimental evaluations, we further suggest experiments based on the extended desiderata listed in \S\ref{desiderata}. The first tests the ability of the model to detect a change in task, which has not yet been demonstrated in continual learning. The second evaluates time against accuracy (or some other metric) as a trade-off that must be carefully selected for different settings.

\subsubsection{Model Uncertainty on Unseen Tasks}\label{mutual_information}
\begin{figure}
\centering
\begin{minipage}{\columnwidth}
\vspace{-2mm}
\includegraphics{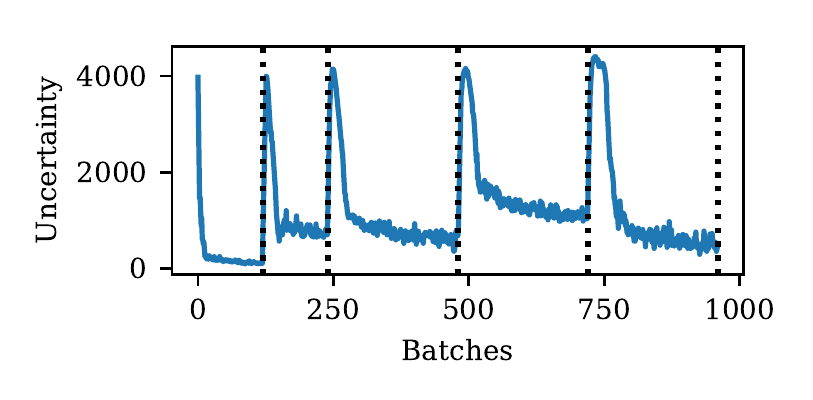}
\vspace{-9mm}
\caption{\textbf{Mutual Information on FashionMNIST with VGR.} Uncertainty spikes at the start of each new task (black dotted line), allowing task boundary detection. Uncertainty is assessed using the mutual information between predictions on each task and the model posterior.} \label{fig:fashion_vgr_mi}
\end{minipage}
\begin{minipage}{\columnwidth}
\vspace{1mm}
\includegraphics[width=\columnwidth]{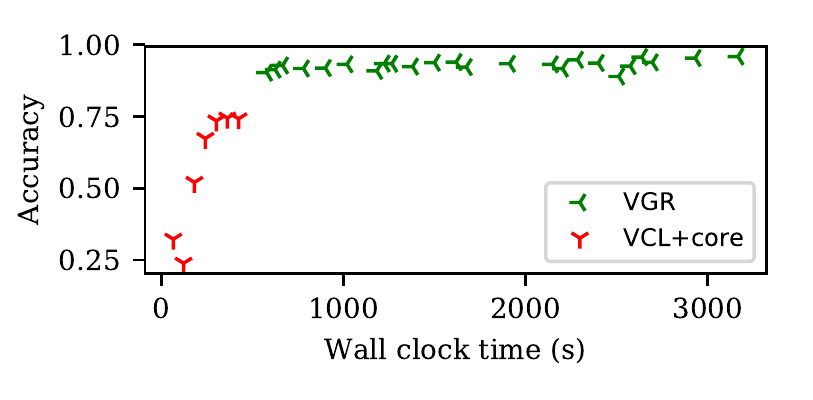}
\vspace{-9mm}
\caption{\textbf{Timed Single-headed Split MNIST.} Hyperparameter choice can lead to slower, more accurate, models for each architecture. Ideal performance is in the top left corner. VGR is accurate but slower than VCL.}
\label{fig:split_second_single}
\vspace{-7mm}
\end{minipage}
\end{figure}

Model uncertainty gives another tool for understanding forgetting. A good continual learning model should be more certain on tasks it has seen before than unseen tasks. To test the ability of model uncertainty to distinguish a task boundary, we assess the uncertainty of the model, for each batch of training, on data from the current task. We find, for a model trained using VGR, that model uncertainty spikes at the start of each new task (see figure \ref{fig:fashion_vgr_mi}). This means that VGR, or another method with sufficiently good model uncertainty estimation, does not need to be told task boundaries. Before each training epoch, we can compare the model uncertainty on the new data to the previous running average uncertainty. If the change in model uncertainty is above any sensible cut-off threshold we see a task boundary has occurred and trigger training of the generative model with a cache of the previous epoch's data. One could either cache the last epoch's data or train the generative model alongside the task training, but either way task detection comes with some additional memory and computational overheads. To measure uncertainty, we use the mutual information between predictions on each task and the model posterior, following \citet{Gal2016}.

\subsubsection{Time and Memory}\label{time}
\textit{Timed Split MNIST} evaluates the trade-off of training time against accuracy in continual learning, rather than reporting a single run-time. Much of the need for continual learning comes from situations where a system cannot feasibly retrain on all previous data due to constraints of cost and time. Architecture selection and hyperparameter selection, therefore involve a tradeoff of performance and speed. For example, in VGR, training on larger sampled datasets improves performance but takes more time. In figure \ref{fig:split_second_single} we show the average accuracy of various models plotted against the wall-clock time taken to finish training for a wide range of such hyper-parameters. While it is good to report the time taken by the most accurate model from one approach, representing the trade-offs within each approach gives a fairer reflection of the costs and benefits of each method. None of the configurations of VCL provides good accuracy on the single-headed task. At the same time, none of the configurations of VGR is able to provide good time performance (see Appendix \ref{comparing_prior_likelihood} for further details and hyperparameter choices).

\section{Discussion}
Experimental design shapes the field by providing reproducible comparisons between architectures, and testing whether an architecture satisfies fundamental objectives. Accidental blindspots in existing evaluations have created misleading comparisons in recent research. When experiments better reflect the continual learning problem, recent leading approaches fail even on simple datasets like MNIST. Using more realistic datasets is important, but until we improve the experimental design, the field will only show illusory progress on those datasets, without overcoming fundamental challenges inherent to continual learning itself. Authors of new papers currently feel compelled by the reviewing process to assess new methods on the same, incomplete, evaluations used by old papers. This paper reflects an attempt to reassess the experimental foundations of continual learning and provide a new direction for future work.

\bibliography{zotero,references}
\bibliographystyle{plainnat}

\newpage

\appendix
\appendixpage
\section{Further Prior Work}\label{furtherpriorwork}

\subsection{Elastic Weight Consolidation}
Unlike VCL, EWC uses an ordinary neural networks with an $L_2$ regularization term added to the loss. This regularization reflects a Gaussian prior for each parameter. At the end of training on each dataset, EWC \citep{kirkpatrick_overcoming_2017} estimates the contribution each parameter makes to the gradient of the loss by approximating the Hessian of the likelihoods with:
\begin{equation*}
F_t \propto \sum^{N_t}_{n=1} \Big( \nabla_{\omega} \mathrm{log}p(y_t^{(n)}|\omega, \mathbf{x}_t^{(n)}) \Big)^2\bigg\vert_{\omega = \omega_{t}} 
\end{equation*}
This approximate Fisher information matrix is estimated at the end of each task. This implicitly sets the covariance of the parameter prior (which is assumed to be diagonal). The authors state that in the multi-task case the regularization term for task $T$ is the sum of separate regularization terms for each past dataset $\mathcal{D}_t$.

\citet{Chaudhry2018} provide a generalization of both SI and EWC. \citet{Ritter2018} extend EWC by relaxing the assumption that the covariance matrix of the parameter distributions is diagonal.

\citet{kirkpatrick_overcoming_2017} show EWC works well on the Permuted MNIST task introduced by \citet{Goodfellow2013}. They also show that EWC reduces forgetting on a succession of Atari games.

\subsection{Synaptic Intelligence} Published soon after EWC, SI has a similar loss but computes the contribution of each parameter to the gradient of the loss over the entire course of training. The authors explicitly acknowledge that the derivation of their regularization term only extends to the two-task case, but point out that the model performs well even in multi-task settings.

\citet{Zenke2017} show that SI works in a \textit{multi-headed} version of the Split MNIST task which they introduce, and matches EWC on the Permuted MNIST task. They also show some transfer learning on a multi-headed split CIFAR task of their own design.

\subsection{Variational Continual Learning}\label{vcl}
\citet{nguyen_variational_2018} motivate their regularization through VI. They argue that the posterior of the parameter distribution after learning a task should be the prior when learning the next task. They calculate the Kullback-Leibler (KL) divergence between the current distribution and the previous posterior. This is in contrast with the more usual Variational Free Energy (VFE) loss which uses the KL divergence to a constant prior that is not updated after each task. The KL divergence can be represented approximately as a quadratic regularization with rotation, just like the two approaches above, as \citet{nguyen_variational_2018} point out.

In addition to the Variational Continual Learning (VCL) regularization term, \citet{nguyen_variational_2018} add a coreset which samples examples from old datasets, which are withheld from the main dataset. Rather than discarding all of an old dataset, as EWC and SI do, only the vast majority (about 99.7\%) is discarded. Before being evaluated, the model is trained against the coreset. For multi-headed settings (see \S\ref{crit_split}) coresets train each head separately.

\citet{nguyen_variational_2018} show that VCL with or without the use of coresets outperforms both SI and EWC on the Permuted MNIST task. They show that VCL outperforms EWC and SI on the multi-headed Split MNIST task and on a very similar task with notMNIST. They take advantage of the VI setting to show reduced uncertainty in a generative task based on MNIST and notMNIST.

\subsection{Variational Generative Replay}\label{vgr}
Each of VCL, EWC, and SI effectively set the parameters at the end of the last task as a prior, which is explicit in the Variational Inference (VI) derivation supporting VCL \citep{nguyen_variational_2018}. Instead of changing priors between tasks, VGR adapts the log-likelihood component of the loss to depend on past datasets \citep{farquhar_unifying_2018}. VGR estimates the log-likelihood component of the loss on past tasks using generative models. They train a DC-GAN at the end of each task on all of the classes that appeared in that task. Then, at the start of training a new task, they sample from all stored GANs training points, which are then mixed into the real training data.

\subsection{Dynamic architecture approaches}
Other work, which is not our focus, has also made progress with continual learning. Progressive neural networks \citep{Rusu2016} conditions the newest model on the outputs of old, stored, models. Some approaches split the network into regions that have different roles. In \citet{Li2017} one set of shared parameters governs a feature extraction component while each task is given parameters on top of that component. During training, the combined shared and old task networks are encouraged to classify similarly to a stored old version of the model, while the combined shared and new task networks are trained on the new task. \citet{Jung2016a} stochastically freeze the final layer to encourage lower layers to extract features from all tasks that the final layer can use to classify. Others have semi- or fully-fixed a shared part of a network with task-specific layers on top \citep{Razavian2014, Yosinski2014a, Donahue2014}. PathNet uses evolutionary selection to try to learn which patches of a larger network are helpful to each task rather than making a layer-by-layer assumption \citep{Fernando2017a}. This broad family of approaches are especially useful in two-task transfer cases, but it can become impractical to introduce the many task-specific weights in the more general continual learning case that we consider.

Others find that adjustments to learning dynamics can reduce catastrophic forgetting, presumably by encouraging networks to use their capacity fully. This family of approaches includes selecting activations and using dropout to reduce forgetting \citep{Srivastava2013a, Goodfellow2013}. These are useful, but do not go far enough to solve forgetting fully. \citet{Lee2017} extend dropout by shifting the zero-point of each dropout parameter to the parameter value from training on the previous task.

\section{Experimental Details and Further Figures}

\begin{figure}
\centering
\begin{tabular}{lc}
\hline 
 & \textbf{ENTROPY} \\ 
\hline 
Split & $0.003$ \\ 
Permuted & $0.453$ \\
\hline
\end{tabular} 
\caption{Average entropy of predictions on Task B, early in training; Note the 2 orders of magnitude difference between the two settings. Entropy is much higher in the Permuted setting.}
\label{table:entropy}

\end{figure}

\subsection{Shortcomings of Permuted MNIST}\label{permuted_problems}
Although our work mostly aims to show that architectures which succeed on a Permuted MNIST task can fail in slightly different settings, we also carried out some investigations as to why this might be. We observe that in most settings, a model will confidently predict that an example from the new dataset $\mathcal{D}_{t}$ is from a class that was in  $\mathcal{D}_{t-1}$. This confident but false prediction creates large gradients in the output layer. The derivative of the likelihood term in the loss with respect to each output weight $w_k$ in the output layer of a model is $\frac{\partial \mathcal{L}}{\partial o_k} = p_k - y_k$ where $y_k$ is the $k$'th entry of a one-hot vector of labels and $p_k$ is the probability predicted for class $k$. This gradient is therefore biggest for confident but false predictions. In the permuted setting, however, no example from $\mathcal{D}_{t}$ looks remotely like an example from $\mathcal{D}_{t-1}$. This means the model makes unconfident predictions and the gradients stemming from the likelihood term of the loss are unusually small. This may on its own be enough to explain part of the difference in the settings.

We may be able to probe more deeply. For prior-focused continual learning, the loss function is separated into a prior term and a likelihood term (c.f. eq. (\ref{eq:prior})). By examining this separation, we can see why the permuted setting, with unusually small likelihood-term gradients, is a best case scenario for prior-focused continual learning. The prior term is \textit{data-independent} and has the same magnitude no matter how incorrectly confident the model is when making predictions on $\mathcal{D}_{t}$. The likelihood term of the gradient, however, has a much larger magnitude when confident but false predictions are made. This means that in a typical setting the likelihood term of the loss dominates the prior early in training. In the permuted setting, however, the prior term is relevant throughout the training. This means that the prior-focused approaches are able to succeed in a permuted setting but fail in other settings. Two points are worth noting. First, if the prior term were to fully the capture correct class of models, its size will change appropriately, and this effect would not exist. But in practice our priors are restrictive approximations. Second, the gradients in earlier layers are less straightforward to analyse. However, especially early in training, output layer gradients are several orders of magnitude larger than gradients in early layers. Moreover, our argument only depends on showing that the Permuted setting protects the prior-focused loss from systematic large mis-estimates for an important subset of its gradients.

We introduce two tests to show the shortcomings of the Permuted MNIST experiment discussed in \S\ref{crit_permuted}. For each test we consider both the Permuted and Split settings. In each case, we train a model on Task A before training on Task B. In the split setting, Task A is the first five digits of MNIST and Task B the last five digits. In the permuted setting, Task A is MNIST and Task B is a random permutation of the pixels of MNIST. Here, we are assessing the evaluation framework itself, not continual learning performance.

\subsubsection{Verifying Permuted Setting Under-Confidence}
We first validate our hypothesis that predictions are much more uniform in the Permuted setting. We evaluate the entropy of the output probability vector. Low entropy indicates a confident prediction of a single class, whereas a high entropy indicates a `uniform' prediction, spreading the mass across different classes. These are summarised in Table \ref{table:entropy}, supporting our claim that models in the Permuted setting are much less confident than in a more typical setting. 

\begin{figure}
\centering
\includegraphics[width=0.95\columnwidth]{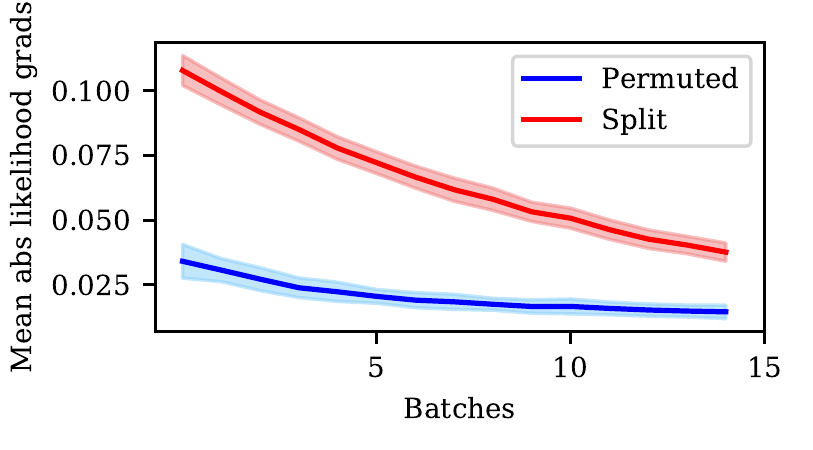}
\vspace{-4mm}
\caption{Early in training Task B, the likelihood term of gradients on the final layer is unusually low in the Permuted setting because permuted digits do not resemble any digits from Task A. This makes continual learning unrealistically easy in this evaluation. Averaged over 100 runs, shading is one standard deviation.}
\label{fig:perm_prob}
\end{figure}

\begin{figure}
\centering
\includegraphics[width=0.95\columnwidth]{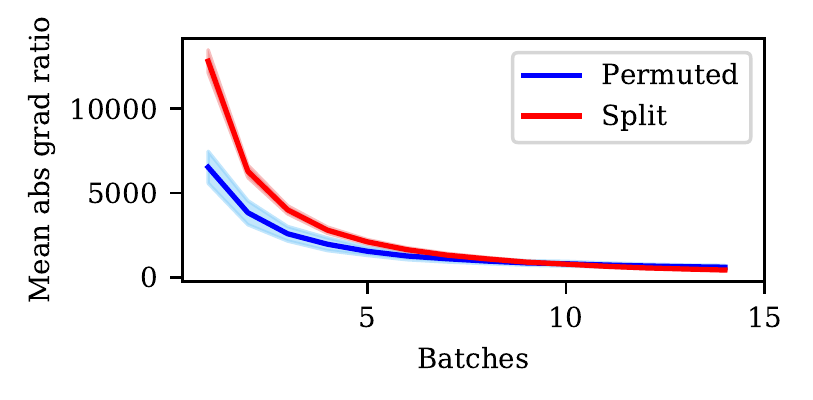}
\vspace{-2mm}
\caption{Early in training Task B, the ratio of the likelihood term of gradients the final layer against the prior term is much lower in the Permuted setting. This reduces forgetting because the prior has a larger impact on learning. Averaged over 100 runs, shading is one standard deviation.}
\label{fig:perm_prob_2}
\end{figure}

\subsubsection{Verifying Unusually Large Gradients in the Permuted Setting}
Next, we test our hypothesis from \S\ref{crit_permuted} that more uniform predictions lead to slower `catastrophic forgetting'.
To test this hypothesis, after training on Task A, we measured gradients for the final weight layer while training Task B in the two variants (figure \ref{fig:perm_prob}). All results are averaged over 100 runs. Full experimental details are in Appendix \ref{permuted_shortcomings_settings}. We found that having digits resembling previously observed ones in the Split setting led to much larger likelihood-term gradients in the final layer than found in the Permuted setting for early batches. As a result, the ratio between the likelihood term and the prior term of the loss is much higher in the Split setting (figure \ref{fig:perm_prob_2}). This means that the prior has a smaller influence over the gradients, leading to more forgetting.

\subsubsection{Experimental Settings for the Above Gradient Analysis}\label{permuted_shortcomings_settings}
To measure the gradients during training, we first trained on Task A for 120 epochs with batch size 256 and an Adam optimizer using Variational Continual Learning (VCL) with the same settings used in \ref{perm_details}. We then trained on Task B for 15 batches of 16 digits each, again using VCL. We measured the average absolute value of the gradients of the final layer of the model. We averaged over 100 training runs for each setting (with a different permutation each time in the Permuted setting), resetting the model to its initial position after each run. Graphs show the standard deviation of the average gradient of each batch. The ratio displayed in figure \ref{fig:perm_prob_2} is the log-likelihood term of the loss described in equation \ref{eq:prior} divided by the prior term, using VCL. The ratio is calculated for each of the 100 runs on the averaged absolute gradients.

\subsubsection{Investigating the Hypothesis that the Prior Term is Overly Small}
In order to investigate whether the prior-term of the gradient is overly small in the Split setting relative to the Permuted setting, we tried training a VCL model on Single-headed Split MNIST with an artificially upweighted prior-term in the cost. We experimented with factors of 1 (the default case), 10, 100, and 1000. We also tried upweighting the prior-term only during the first few epochs. In no case did we find that this improved performance significantly. This suggests that it is not only the magnitude, but also the direction, of the prior gradient which is inaccurate in the split setting.

\begin{figure}
\includegraphics[width=\columnwidth]{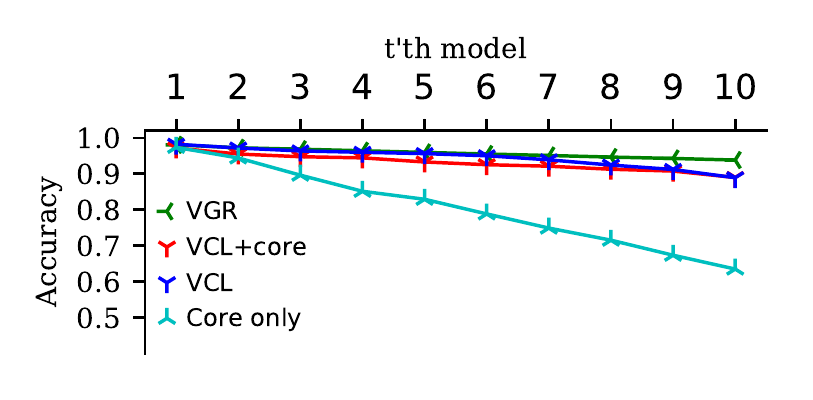}
\caption{\textbf{Permuted MNIST.} VGR has marginally better performance than VCL with or without a coreset, which has been shown to outperform EWC and SI. Coresets without the VCL loss performs worse. Averaged over 10 runs.}
\label{fig:permuted_avg_vgr}
\end{figure}

\subsection{Permuted MNIST as Baseline}\label{perm_details}
As \citet{nguyen_variational_2018} have shown, on Permuted MNIST, described in \S\ref{crit_permuted}, VCL performs well with or without a coreset. We further show that the coreset on its own performs badly (figure \ref{fig:permuted_avg_vgr}, results averaged over 10 runs). This is consistent with our analysis of the Permuted setting, which suggested that the small and evenly spread gradients due to the likelihood term of the loss helped the prior term to work, despite a prior that does not fully support the function space.  Note that this version of VCL was previously shown to outperform EWC and SI \citep{nguyen_variational_2018}.

For permuted MNIST, we follow \citet{nguyen_variational_2018} where possible. We use a Bayesian neural network \citep{Graves2011} with two hidden layers of 100 units with ReLu activations. The priors are initialized as a unit Gaussian and the parameters are initialized with the mean of a pre-trained maximum likelihood model and a small initial variance ($10^{-6}$). We use the Adam optimizer \citep{Kingma2015} with learning rate $10^{-3}$. We use a single head, again following \citet{nguyen_variational_2018}. For all results we present the average over 10 runs, with a different permutation each time. Standard errors are not shown as they are under a tenth of a percent.

We train for 100 epochs using a batch size of 256 on all the data except the with-held coresets. We train the whole single head on coresets for 100 epochs and use 200 digits of each permutation as a coreset chosen using the same k-center coresets algorithm used by \citet{nguyen_variational_2018}. VCL without coresets is exactly the same, but without a final training step on coresets. The coresets only algorithm is exactly the same as VCL, except that the prior is always initialized as though it were the first task.

Following \citet{farquhar_unifying_2018} where possible, we train VGR for 120 epochs using a GAN trained for 200 epochs on each MNIST digit. We use 6000 generated digits per class, sampled fresh for each task, and initialize network weights using the previous task. We use a batch size of 256 times the number of seen tasks, ensuring that the number of batches is held constant. The GAN is trained with an Adam optimizer with learning rate $2*10^{-4}$ and $\beta_1$ of $0.5$. The network has four fully-connected hidden layers with 256, 512, 1024 and 784 weights respectively. It uses Leaky ReLu with $\alpha$ of 0.2.

\subsection{Experimental Settings for Single-headed Split MNIST}\label{single_details}
The settings for Split MNIST follow \citet{nguyen_variational_2018} where possible. For Bayesian neural network architectures, we use exactly the same settings as for Permuted MNIST, including the single head, except that each hidden layer has 256 weights, similarly to \citet{nguyen_variational_2018} on their multi-headed Split MNIST. Results are shown averaged over 10 runs, with a different coreset selection each time. Standard errors are not shown as they are of the order of a tenth of a percent.

For all Bayesian neural network architectures we train for 120 epochs. We use batch sizes equal to the training set size. We use coresets of digits per task selected using the same k-center coreset algorithm as \citet{nguyen_variational_2018}, which are withheld from the training. We train for 120 epochs on the concatenated coreset across all the heads together.

Following \citet{farquhar_unifying_2018} where possible, for VGR we use 6000 digits per class generated by a convolutional GAN. Unlike VCL, we cap batch sizes at 30,000 rather than having the batch size equal the training set size. The GAN is trained for 50 epochs on each MNIST class using the same optimizer as the non-convolutional GAN used for Permuted MNIST. It has a fully connected layer followed by two convolutional layers with 64 and 1 channel(s) and 5x5 convolutions. Each convolutional layer is preceded by a 2x2 up-sampling layer. The activations are Leaky ReLu's with $\alpha$ of $0.2$.

For EWC, we use a neural network with two hidden layers with 256 weights. We train using the variant of EWC suggested in \citet{huszar_note_2018}. We estimate the Fisher information using 200 samples at the end of each task, train using SGD with learning rate $10^{-2}$ for 20 epochs per task. The coefficient of the regularization term in the loss was set to 10 but was found not to be important to performance.

\subsection{Experimental Settings for Multi-headed Split MNIST}\label{multi_details}
Multi-headed Split MNIST has almost precisely the same settings as the Single-headed Split MNIST above. Following \citet{nguyen_variational_2018}, for VCL we have five heads in the final layer, each of which is trained separately. Coresets are used to train each head in turn. Batch size equal is to training set size. Results are shown averaged over 10 runs with fresh coreset selection each time. Standard errors are not shown as they are of the order of a tenth of a percent.

EWC is configured in exactly the same way as in the single-headed setting.

\subsection{Experimental Settings for Split Fashion MNIST}\label{fashion_settings}
Single- and multi-headed Split Fashion MNIST are performed mostly the same as our Split MNIST experiments. We use a larger network, with four hidden layers with 200 units for the BNN and 256 units for EWC. Fashion MNIST has a much more diverse membership of its classes, which makes performance lower even with these larger networks. We deliberately did not optimize hyperparameters, given the observation of \citet{pfulb_comprehensive_2018} concerning causality. However, once hyperparameters were set, we explored somewhat and found that using larger coresets and networks could improved performance slightly, but not dramatically. For VGR, the DC-GAN followed the implementation at https://github.com/carpedm20/DCGAN-tensorflow

\subsection{Split MNIST with Retraining}\label{retraining}
\begin{figure}
\centering
\includegraphics{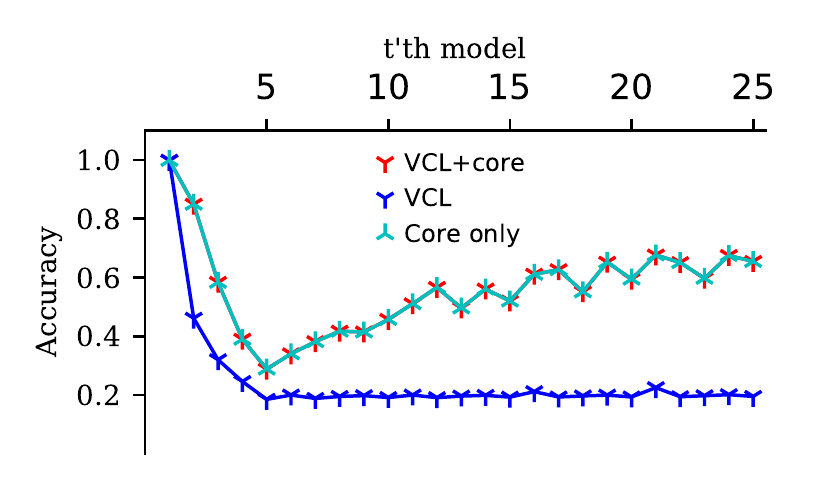}
\caption{\textbf{Retraining MNIST.} With retraining, single-headed Split MNIST is still able to differentiate likelihood-focused and prior-focused approaches.} \label{fig:retraining}
\end{figure}

To test the impact of retraining, we allowed ourselves to cycle through all the data 5 times, which is roughly what previous authors who have made use of retraining have done. Instead of finishing at the end of the fifth task, we begin fresh (keeping all coresets if applicable). We find that even with retraining, single-headed Split MNIST is still able to differentiate prior- and likelihood-focused approaches (see figure \ref{fig:retraining}). This suggests that experiments with retraining may not be misleading in that they make models that fail look as though they work, though it certainly conflicts with some of the use cases that motivate continual learning. For these experiments, we reduced the number of epochs trained per task from 120 to 10, but other settings are the same.

\subsection{Timed Single-headed Split MNIST}\label{timed_split_mnist}
We perform the single-headed Split MNIST experiment with a range of different configurations. For VGR we allow between 10 and 50 epochs for training the convolutional GANs and between 1 and 120 epochs for training the main model. We use 2000, 4000, or 6000 generated images per class. We add GAN training time to the main figure, even though it is possible to do in parallel. For VCL, we use between 1 and 120 epochs for training and coresets. In all cases, we report elapsed wall-clock time from start to finish. We plot this against average accuracy over all tasks of the final model trained on all tasks. In all cases, the experiment was carried out on an NVIDIA Tesla K80.

\begin{figure}
\centering
\includegraphics{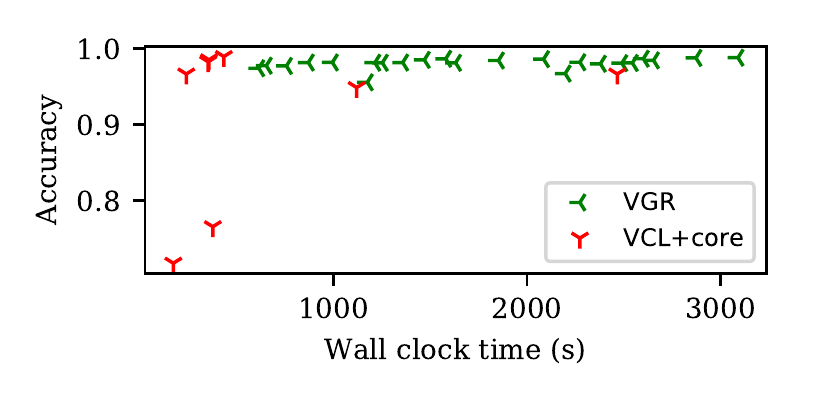}
\caption{\textbf{Multi-headed Timed Split MNIST.} Ideal performance is in the top left corner. VGR can reach good accuracy, but slower than VCL. Now that VCL can use multi-headed training as well as coresets, the accuracy is good in the multi-headed experiment.} \label{split_second_MNISTmulti}
\end{figure}

Here, we also show Timed Split MNIST performed as a multi-headed experiment in order to allow VCL to use multi-headed training (figure \ref{split_second_MNISTmulti}). Performance is better for all models than in the single-headed case, but for VCL it can get very good, and faster than VGR.

\subsubsection{Time and Memory efficiency of Prior- and Likelihood-focused Approaches}\label{comparing_prior_likelihood}
Likelihood-focused approaches must train generative models and train on extra data relative to prior-focused approaches. But prior-focused approaches tend have slightly more complicated losses or add extra training phases. We found that VCL with coreset was faster than VGR. For the single-headed Split MNIST task it took roughly 7 minutes.\footnote{Times are reported for a Nvidia Tesla K80.} Training VGR with each generated set the same size as the true data took roughly 18 minutes plus 34 minutes to train the GANs. But a quicker training regime took only 2 minutes plus 8 for training GANs and was still more accurate than VCL. \footnote{Average single-headed accuracy of the final model for this less-trained VGR was 90\% rather than 96\% for the fully trained model. Training used one third the generated data, only 20 epochs of training, and only a fifth the training for the GAN.} The prior-focused costs scale badly with model size, while VGR's costs scale badly with data-space complexity and very long series of datasets. VGR is also potentially less memory intensive. Generated data can be sampled on demand, it need not be the same each epoch.

\subsection{A Simpler Model Uncertainty Experiment}
\begin{figure}
\includegraphics[width=0.7\columnwidth, keepaspectratio]{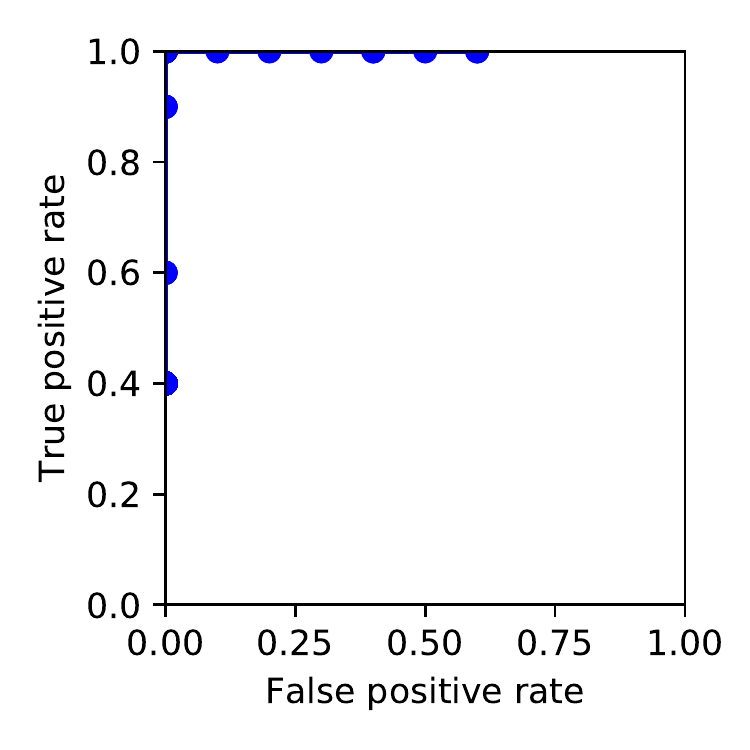}
\caption{For VGR, sensible cut-off points cleanly differentiate previously-seen tasks from unseen ones with no false positives. This results in perfect recognition of previously seen tasks.}
\label{MIplotVGER}
\end{figure}

\begin{figure}
\includegraphics[width=0.7\columnwidth, keepaspectratio]{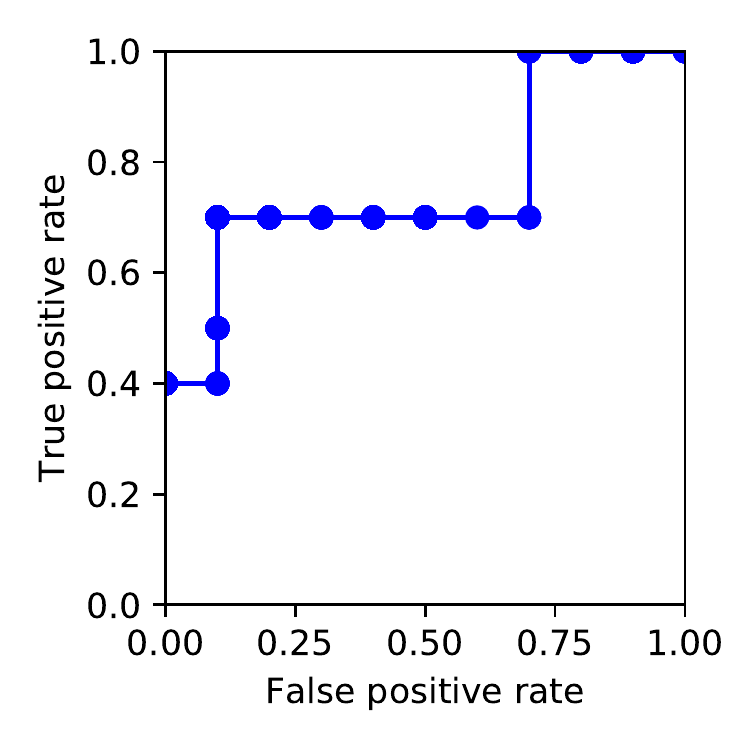}
\caption{For VCL, false positives are a problem because the model is highly uncertain about many previously-seen tasks. This represents forgetting of old tasks.}
\label{MIplotVCL}
\end{figure}

It is possible to measure the quality of model uncertainty for continual learning without tracking throughout training. After training on each task, we measure the model's uncertainty when shown data from each task, both seen and unseen (see figures \ref{MIplotVGER} and \ref{MIplotVCL}).  We then set an uncertainty threshold, when the uncertainty is above this value we predict that the data was previously unseen. By varying thresholds we generate an ROC plot. The area under the curve (AUC) of the ROC plot is a measure for the ability of the model to distinguish seen/unseen tasks. We compare the AUC of both these ROC plots and find that VGR's AUC is 1 while VCL's is 0.76. This means that VGR is in principle able to correctly detect all tasks it has not seen before, whereas VCL fails on a considerable number of those.

\begin{figure}
\includegraphics{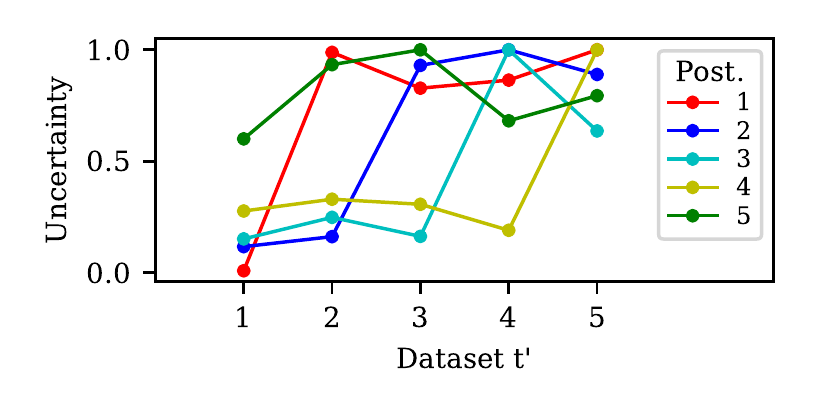}
\caption{\textbf{Mutual Information MNIST VGR.} Uncertainty is high for unseen tasks, and low for all previously seen tasks, showing good calibration. Models trained on 1-5 tasks are each evaluated against all five tasks. Uncertainty is assessed using the mutual information between predictions on each task and the model posterior. It is normalized to 1 for the most uncertain task for each model.} \label{fig:mi}
\end{figure}
\begin{figure}
\includegraphics{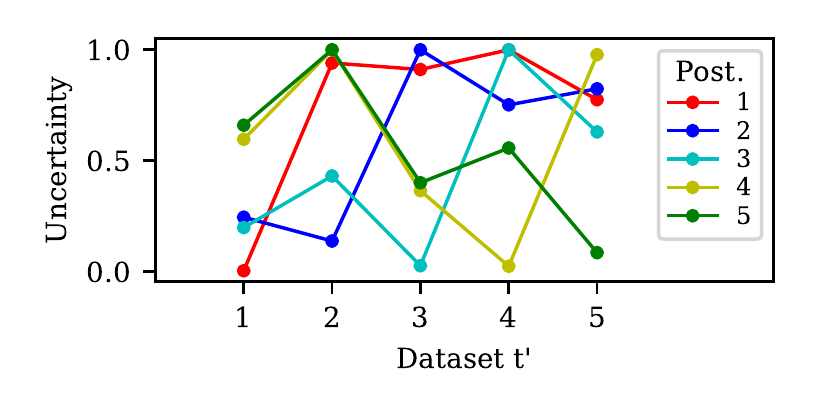}
\caption{\textbf{Mutual Information MNIST VCL.} Uncertainty is high for unseen tasks, but rises for seen tasks that were seen some time ago. The fourth model, for example, is nearly as uncertain about the second task, which it has seen, as it is about the fifth, which it has not.} \label{fig:miVCL}
\end{figure}


\end{document}